\documentclass[10pt,twocolumn]{article}

\usepackage[margin=0.75in,top=0.9in,bottom=0.9in]{geometry}
\usepackage{mathptmx}
\usepackage[T1]{fontenc}
\usepackage[utf8]{inputenc}
\usepackage{microtype}
\usepackage{graphicx}
\usepackage{booktabs}
\usepackage{multirow}
\usepackage{array}
\usepackage{tabularx}
\usepackage{caption}
\usepackage{subcaption}
\usepackage{flushend}
\usepackage{dblfloatfix}
\usepackage{float}
\usepackage{placeins}
\usepackage{hyperref}
\usepackage{xcolor}
\usepackage{amsmath}
\usepackage{natbib}
\usepackage{titlesec}
\usepackage{abstract}
\usepackage{url}

\titleformat{\section}{\normalfont\large\bfseries}{\thesection.}{0.5em}{}
\titleformat{\subsection}{\normalfont\normalsize\bfseries}{\thesubsection}{0.5em}{}
\titleformat{\subsubsection}{\normalfont\normalsize\itshape}{\thesubsubsection}{0.5em}{}
\titlespacing{\section}{0pt}{8pt}{4pt}
\titlespacing{\subsection}{0pt}{6pt}{3pt}
\titlespacing{\subsubsection}{0pt}{4pt}{2pt}

\hypersetup{colorlinks=true,linkcolor=black,citecolor=black,urlcolor=black}

\newcolumntype{L}[1]{>{\raggedright\arraybackslash}p{#1}}
\newcolumntype{C}[1]{>{\centering\arraybackslash}p{#1}}

\begin{document}

\twocolumn[{%
\begin{@twocolumnfalse}
\begin{center}
{\LARGE\bfseries Can Large Language Models Revolutionize Survey Research?\\[4pt]
Experiments with Disaster Preparedness Responses}
\vspace{10pt}

{\normalsize Yan Wang$^{a*}$,\quad Ziyi Guo$^{b}$,\quad Christopher McCarty$^{c}$}
\vspace{6pt}

{\small
$^{a}$Dept.\ of Urban and Regional Planning \& Florida Institute for Built Environment Resilience,
University of Florida, P.O.\ Box 115706, Gainesville, FL 32611.\ \texttt{yanw@ufl.edu} \textsuperscript{*}Corresponding author\\[2pt]
$^{b}$Dept.\ of Urban and Regional Planning \& Florida Institute for Built Environment Resilience,
University of Florida, 1480 Inner Rd., Gainesville, FL 32611.\ \texttt{ziyiguo@ufl.edu}\\[2pt]
$^{c}$College of Liberal Arts and Sciences, Bureau of Economic and Business Research,
University of Florida, Suite 150, 720 SW 2nd Ave., Gainesville, FL 32611.\
\texttt{ufchris@ufl.edu}}
\end{center}
\vspace{8pt}

\begin{abstract}
Survey research faces mounting structural challenges: declining response rates, sample bias, block-wise missingness concentrated among the most at-risk respondents, and a growing wave of AI-assisted fraudulent completions in online panels. Large language models (LLMs) have been proposed as a remedy, yet rigorous empirical evaluations of their performance across the full survey workflow remain scarce, particularly in disaster contexts where data quality and population coverage matter most. We present and evaluate a five-stage framework for LLM integration in survey research covering questionnaire design, sample selection, pilot testing, missing-data imputation, and post-collection analysis, using the 2024 Hurricane Milton preparedness survey of Florida residents ($n=946$) as a shared empirical testbed throughout \citep{wang2026}. We introduce a Protection Motivation Theory (PMT)-constrained co-occurrence knowledge graph and use it to develop seven LLM configurations spanning zero-shot inference, retrieval-augmented baselines, and novel theory-informed variants. Our proposed Anchored Marginal Theory-Informed LLM (A-TLM) outperforms all three classical imputation baselines (IPW/MI, MICE+PMM, and missForest) on root-mean-square error under disaster-relevant block-wise missing-not-at-random conditions (S4 RMSE 1.439 vs.\ 1.496 for the next-best method), while achieving near-zero overall signed bias ($-0.121$) where the random-forest imputer produces the largest absolute bias ($-0.631$). Organizing retrieval around PMT's causal structure and integrating all evidence in a single model call outperforms both unstructured nearest-neighbor retrieval and staged sequential inference (Marginal-TLM combined MAE 0.993 vs.\ 1.097 for standard RAG). We document that near-zero aggregate bias can mask opposing subgroup errors of substantial magnitude, with compound-vulnerable respondents systematically under-predicted across LLM configurations, and propose subgroup-stratified bias auditing as a reporting standard for policy-relevant LLM-augmented workflows. A retrieval-constrained knowledge-graph chatbot demonstrates that hallucination risk is architecturally manageable through grounded refusal. Across all five stages, we identify specific boundaries on LLM utility and provide reproducible per-stage scripts anchored to real disaster survey data.

\smallskip\noindent\textbf{Keywords:} generative AI; natural language processing; surveys; retrieval-augmented generation; in-context learning; disaster response; missing data imputation; Protection Motivation Theory.
\end{abstract}
\vspace{10pt}
\end{@twocolumnfalse}
}]

\section{Introduction}

Survey research stands as a cornerstone of the social sciences, gauging public attitudes, uncovering patterns in human behavior, and shaping policy. However, the field is confronted by escalating methodological challenges, including frequent low response rates, sample bias that fails to represent the target population, and the proliferation of disengaged or fraudulent answers. These challenges have intensified as survey modes have shifted from traditional telephone interviews to self-administered online panels \citep{dillman2005,stern2014,wang2004}. Historically, approaches such as post-stratification weighting and demographic imputation, often guided by benchmarks like the American Community Survey (ACS), have aimed to preserve representativeness and correct for nonresponse. Yet these techniques are increasingly limited in their ability to address partial responses, shifting population dynamics, and complex patterns of missingness.

Compounding these difficulties, reliance on commercial survey panels introduces further opacity: recruitment methods and sample maintenance are often undisclosed, and resulting panels can diverge substantially from their intended populations \citep{bentley2020,callegaro2014,hays2015}. Reviews of consumer sentiment surveys have highlighted how incentives and self-paced online panels can unintentionally inflate results and introduce quality risks that are difficult to trace or correct \citep{hacıoglu2025}. Survey data quality is also compromised by insincere or careless respondents who may provide hasty, random, or fabricated answers \citep{meade2012}.

Recent advances in large language models (LLMs) \citep{zhao2025} have prompted researchers to reconsider longstanding limitations in survey methodology \citep{arora2025,jung2025,perc2025,qu2024,salecha2024}. LLMs offer the potential to augment or streamline nearly every stage of the survey research process \citep{chakraborty2025}: dynamic questionnaire development, adaptive translations for diverse subpopulations, real-time detection and intervention for disengaged respondents, and sophisticated synthesis of both quantitative and qualitative data (see Table~\ref{tab:framework}). With the ability to process and cluster thousands of open-ended responses, LLMs present new opportunities to uncover latent patterns in attitudes and behaviors, going beyond traditional variable-by-variable analysis.

Yet integrating LLMs into survey workflows is not without risk. These models can inherit or amplify biases present in their training corpora, raising concerns about fairness and representativeness, especially when deployed in culturally diverse contexts \citep{ashwin2025,gao2025}. Data privacy is an ongoing concern, as is the possibility of ``hallucination,'' meaning plausible but misleading or entirely incorrect model output \citep{zhao2025}. Moreover, practical and ethical constraints in model deployment, supervision, and interpretation challenge their scalability, reinforcing the need for vigilant human oversight at every step \citep{perc2025}.

The stakes are especially high in \textbf{disaster survey research}, where methodological challenges are intensified by rapidly changing populations, heightened respondent distress, and the urgent need for timely, actionable information. Pre-disaster studies grapple with uncertain population boundaries, low engagement with hypothetical scenarios, and sparse preparedness data \citep{hao2022}, while post-disaster surveys face disrupted communications, increased respondent distress, and compressed timelines \citep{king2002}. The 2024 Hurricane Milton preparedness survey of Florida residents, which forms the empirical backbone of this study, documents how routine time constraints interact with hurricane preparation behavior across household types \citep{wang2026}. In these demanding contexts, traditional approaches to sampling, data collection, and quality control can quickly become inadequate.

The intersection of LLMs and human expertise in disaster research represents a promising yet underexplored frontier. This study uses disaster preparedness and response as a test case to systematically investigate these questions, drawing on experimental analyses and the Hurricane Milton RAPID survey to identify practical strategies and future directions for effective human-AI collaboration. We empirically evaluate the strengths and pitfalls of LLMs across five operational stages of the survey-research workflow on the 2024 Hurricane Milton disaster-preparedness survey \citep{wang2026}, benchmarking LLM-based imputation against three established classical baselines (IPW/MI, MICE+PMM, and missForest). Our goal is to illuminate both the promise and the limits of LLM-enhanced survey research, providing groundwork for responsible guidelines and future innovation at the intersection of AI and social measurement.

\begin{table*}[!ht]
\centering
\caption{Large Language Models in Survey Research: An Adoption Framework.}
\label{tab:framework}
\small
\begin{tabularx}{\textwidth}{L{2.5cm} X L{3.5cm}}
\toprule
\textbf{Survey Research Stage} & \textbf{How LLMs Can Assist} & \textbf{Role of Human Supervisor / Partner} \\
\midrule
Research \& Questionnaire Design &
Streamline and create survey questionnaires. Adapt and translate for different subpopulations and modes (landlines, cellphone, online). Leverage extensive databanks for thorough item generation. Suggest multidimensional constructs or latent variables. &
Ensure model is trained on relevant data; oversee ethical considerations in instrument development. \\
\addlinespace
Sample Selection &
Model sample characteristics for targeted recruitment. Optimize representativeness across survey modes. Aid prediction of difficult-to-reach samples and inform allocation of resources. Advance beyond opt-in panels as LLM may help subpopulation sampling. &
Guide selection criteria; evaluate representativeness and sampling quality; interpret mode-sensitive sampling implications. \\
\addlinespace
Pilot Testing &
Generate synthetic respondent data; anticipate analytic challenges (e.g., by preparing mock tables and outputs); test survey design robustness prior to fielding. &
Interpret synthetic datasets; refine pilot instruments based on LLM outputs; verify utility for real samples. \\
\addlinespace
Data Collection &
Impute missing data. Generate synthetic responses as needed. Conduct bias mitigation. Identify and flag ``bogus'' or disengaged respondents in real time. Intervene on quality (e.g., mid-survey clarifications or motivational prompts). &
Monitor ongoing data quality; adjudicate intervention strategies; oversee ethical management of respondent flagging and replacement. \\
\addlinespace
Data Analysis &
Extract key themes from open-ended responses; conduct statistical modeling and formal analyses; produce rapid visualizations and summaries. &
Review and interpret analytic outputs; ensure validity of findings; contextualize results for reporting and decision-making. \\
\bottomrule
\end{tabularx}
\end{table*}

\section{Background}

\subsection{Current Challenges in Survey Research}

\subsubsection{Evolution of Survey Research Methods}
Survey methodology has evolved through a sequence of distinct technological eras, each defined by trade-offs among coverage, cost, and data quality. Face-to-face interviewing produced the highest response rates but at prohibitive per-interview cost \citep{groves1998}. The telephone survey era reduced costs substantially while preserving many quality advantages of human contact. The theoretical frameworks developed during these decades remain foundational: Groves and Couper's \citeyearpar{groves1998} leverage-saliency model of respondent cooperation and Tourangeau et al.'s \citeyearpar{tourangeau2000} four-stage model of survey response continue to anchor contemporary item design and mode-effect research.

The transition to self-administered web surveys further disrupted the field's economics while intensifying quality concerns. Response rates fell steeply, and the Total Survey Error (TSE) framework, which decomposes survey estimates into errors of non-observation and observation, became the standard vocabulary for diagnosing these losses \citep{groves2010}. The Tailored Design Method \citep{dillman2014} codified mixed-mode best practices, while Couper's \citeyearpar{couper2017} review documented the full scope of web displacement: collapsing response rates, the proliferation of probability-based and opt-in panel designs, and the integration of passive behavioral data into survey architectures. Recent innovations in mixed-mode design, including sequential web-phone protocols and push-to-web strategies, have recovered some ground, with documented gains of 12 to 25 percentage points over single-mode designs \citep{coffey2024}, but the secular decline in public cooperation shows no sign of reversal. Federal economic surveys have seen response rates fall by approximately 15 to 30 percentage points over the past decade \citep{leduc2025}.

\subsubsection{Data Quality Challenges in Modern Surveys}
Beneath the headline problem of declining response rates lies a more troubling development: the growing unreliability of completed responses. Multiple imputation, formalized by \citet{rubin1987} as the principled replacement of each missing value with $M$ draws from its posterior predictive distribution combined via Rubin's Rules, remains the statistical standard for item-level nonresponse. MICE with predictive mean matching \citep{vanbuuren2011} has become the dominant implementation. Inverse-probability weighting combined with multiple imputation extends coverage to block-level missingness \citep{seaman2012}, and non-parametric random-forest imputers handle mixed-type data with non-linear dependencies \citep{stekhoven2012}. These three methods constitute the classical comparison set for the LLM-based approach evaluated here.

The integrity of completed responses has also come under mounting pressure. \citet{pinzon2024} audited 31 fraud-detection strategies currently deployed in commercial panels and found that none adequately maintain analytic-quality samples, with usable completion rates falling from roughly 75\% to 10\% in some panels between 2021 and 2024. Johnson et al.\ \citeyearpar{johnson2024} report that approximately 40\% of submissions in a participatory-mapping web survey were classified as fraudulent despite multi-layer behavioral and IP screening. \citet{westwood2025} demonstrates that autonomous AI agents can pass standard attitudinal surveys at a 99.8\% success rate against conventional attention checks, and documents that over one-third of Prolific respondents admitted using LLMs to compose open-ended answers. The macroeconomic consequences are already visible: consumer sentiment indicators diverged sharply from verified retail-purchase behavior during 2023 to 2024, attributable in part to mode effects and recruitment-composition shifts in underlying panels \citep{hacıoglu2025}.

\subsection{Potential of Large Language Models}

\subsubsection{Overview of LLMs}
Large language models are transformer-based neural networks trained on massive text corpora to model the conditional probability of the next token given its context. GPT-3 demonstrated that scaling alone yields a general-purpose model capable of competitive performance across diverse natural language tasks using only in-context examples, without task-specific fine-tuning \citep{brown2020}. This finding reoriented the NLP paradigm from task-specific architectures to prompt-conditioned general models, a shift subsequently extended and safety-aligned in frontier systems such as GPT-4 \citep{openai2023}. Comprehensive surveys document this landscape in detail, covering model families, training paradigms, benchmark performance, and open challenges \citep{minaee2025,naveed2025,stanford2025}.

The principal limitation constraining scientific deployment of LLMs is hallucination: the generation of outputs that are syntactically plausible but factually incorrect or internally inconsistent. Huang et al.\ \citeyearpar{huang2024} distinguish factuality hallucination from faithfulness hallucination and document that mitigation strategies must be addressed throughout the entire model lifecycle. Even frontier systems hallucinate on a non-trivial share of prompts in some task categories, with rates varying systematically with prompt structure and retrieval design \citep{ravi2025,saxena2025,frontiers2025}. Retrieval-augmented generation (RAG), which grounds model outputs in explicitly retrieved evidence rather than relying on parametric memory \citep{lewis2020}, provides partial but meaningful mitigation and is the primary architectural strategy adopted in the pipeline evaluated here.

\subsubsection{LLMs and Survey Research}
The most influential early study of LLMs in survey contexts is Argyle et al.'s \citeyearpar{argyle2023} ``silicon sampling'' framework, in which GPT-3 was conditioned on detailed demographic backstories derived from real survey respondents. The model's outputs matched human response distributions and cross-item correlation patterns with what the authors termed high ``algorithmic fidelity.'' Sun et al.\ \citeyearpar{sun2024a} extended this to group-level conditioning, showing that LLM-generated distributions closely approximate real public-opinion marginals for many items while performing worst on socially sensitive topics where models systematically produce what the authors term \textbf{harmlessness bias}, a shift toward socially desirable answers. \citet{horton2023} evaluated LLMs as simulated economic agents and found internally coherent but training-data-anchored behavior, positioning LLMs as instruments for theory exploration rather than human-subject substitutes.

A parallel body of work has catalogued the risks of this approach. \citet{bisbee2024} find that silicon-sampled GPT-3.5 responses inflate estimates of partisan and racial affective polarization by a factor of roughly seven relative to human benchmarks, attributing the distortion to stereotype amplification in persona conditioning. \citet{ashwin2025} demonstrate that LLM-based qualitative coding produces errors systematically correlated with interviewee characteristics such as refugee status, gender, and education level, leading to biased inference. Performance gaps are especially severe in non-Western and non-English contexts, consistent with training-corpus imbalance \citep{sun2024b}. Synthesizing 285 silicon-human comparisons, \citet{sarstedt2024} conclude that LLMs replicate surface-level patterns but fail to reproduce many deeper behavioral regularities, making them most credible at upstream design stages. \citet{crockett2025} argue further that AI surrogates can reinforce existing sampling biases, creating an illusion of generalizability rather than genuinely expanding it.

On the constructive side, AI-assisted conversational interviewing has been shown to enhance the specificity and explanatory depth of open-ended responses, though with trade-offs in respondent burden and attrition \citep{barari2025}. LLM-based phone agents achieve lower item-nonresponse than self-administered web forms while presenting elevated risk of leading-question artifacts \citep{kaiyrbekov2025}. \citet{loru2025} caution against delegating evaluative judgment to models susceptible to what they term \textbf{epistemia}, the illusion of knowledge produced when surface plausibility substitutes for contextual verification. Augmenting LLMs with variational autoencoders to induce individual heterogeneity substantially improves performance on collective-opinion benchmarks \citep[CrowdLLM;][]{lin2024}, and PersonaFuse demonstrates that dynamically adapting personality traits to conversational context opens new possibilities for adaptive survey interviewing \citep{tang2025}. Hybrid LLM-statistical approaches outperform either alone on distributional reconstruction \citep{miranda2025}.

\subsection{Disaster Survey Research: Unique Contexts and Constraints}
Disaster survey research operates under constraints that amplify every methodological challenge described above. Population boundaries are inherently unstable: evacuation, sheltering, displacement, and mortality all alter the sampling frame between hazard onset and the earliest feasible data collection, often in ways that cannot be fully characterized until the survey itself is complete \citep{king2002}. Pre-disaster studies face the additional problem of measuring behavioral intentions toward low-probability, high-consequence events that many respondents have never experienced, generating responses anchored to hypothetical scenarios rather than to revealed behavior \citep{lazo2015}. Post-disaster studies face the inverse: real behavior did occur, but the respondents who experienced it most intensely are those most likely to be missing from the sample due to infrastructure disruption, psychological distress, or temporary displacement \citep{hao2022}. The resulting data frequently exhibit block-wise missingness, where entire question modules are absent for identifiable subgroups, a pattern that standard item-level imputation routines are not designed to address.

Protection Motivation Theory (PMT) provides the dominant explanatory framework for preparedness behavior in this context \citep{floyd2000}. PMT posits that protective action follows from the joint output of threat appraisal, integrating perceived severity, perceived vulnerability, and prior experience, and coping appraisal, integrating response efficacy, self-efficacy, and response cost. In disaster-preparation contexts, response cost has both a financial and a temporal dimension, and for time-constrained households the temporal dimension is often the binding constraint on protective behavior \citep{lazo2015,slovic2007,wang2026}. The Hurricane Milton instrument is specifically designed to index this temporal dimension in depth, providing the theoretical scaffold for the knowledge-graph methods evaluated here.

\section{Data, Survey Instrument, and Theoretical Framework}

\subsection{Data and Survey Instrument}
All operations are anchored in the 2024 Hurricane Milton preparedness survey of Florida residents ($n=946$), collected via online panel during the weeks following landfall and described in detail in \citet{wang2026}. The instrument is organized into three blocks. \textbf{Block A} (always observed; 16 items) contains demographic items including age, gender, race or ethnicity, marital status, education, employment, occupation, income bracket, housing tenure, household size, caregiving burden, dependent age categories, a health-issue indicator, and prior hurricane experience. \textbf{Block B} (routine time use; 8 ordinal items) measures daily time allocation, perceived time scarcity, and schedule flexibility. \textbf{Block C} (hurricane preparation; 8 ordinal items) measures awareness timing, preparation timing and duration, preparation stress, time spent on dependents, and personal disruption.

A deterministic 80/20 split (seed$=42$, by respondent identifier) yielded a training set of 757 records and a validation set of 189 records. We derived four binary vulnerability flags from Block A: low-income households (income at or below \$45,999 USD), minority (race other than ``White''), renter (apartment or mobile home), and disabled (reported health issue other than ``No difficulties experienced''). Respondents satisfying at least two flags were classified as \textbf{compound-vulnerable}. The validation set contained 72 compound-vulnerable respondents (38\%), sufficient for stratified evaluation. Where comparison to population baselines was required, we used the U.S.\ Census Bureau American Community Survey 5-year estimates for Florida adults from vintages 2018 to 2022 \citep{uscensus2024}.

\subsection{Theoretical Framework: Protection Motivation Theory}
PMT specifies the cognitive sequence by which an individual translates a perceived threat into protective action \citep{floyd2000}. The individual first conducts a threat appraisal, estimating perceived severity and perceived vulnerability weighted by prior experience and affective response. The individual then conducts a coping appraisal, estimating response efficacy (whether the action works), self-efficacy (whether one can perform it), and response cost (what the action consumes in time and resources). The differential between the two appraisals produces a protection motivation that, when sufficiently activated, translates into protective behavior. In disaster-preparation contexts, for time-constrained households the temporal dimension of response cost is often the binding constraint \citep{lazo2015,slovic2007,wang2026}.

The Hurricane Milton instrument indexes the temporal dimension of response cost in depth, with 16 outcome items spanning routine time allocation, perceived time constraints, and preparation timing. It does not, however, provide multi-item scales for threat appraisal, coping appraisal, or protection motivation in the conventional PMT sense. We therefore operationalize PMT as a six-stage temporal-cost cascade aligned to the available items (Table~\ref{tab:pmt}), with each construct estimated solely from Block A demographics or from upstream cascade stages, ensuring that no prediction target enters its own conditioning set.

\begin{table*}[!ht]
\centering
\caption{PMT Cascade Operationalized on the Hurricane Milton Instrument.}
\label{tab:pmt}
\small
\begin{tabularx}{\textwidth}{C{0.55cm} L{3.2cm} C{0.9cm} X}
\toprule
\textbf{Stage} & \textbf{PMT Construct} & \textbf{Block} & \textbf{Survey Variables} \\
\midrule
1 & Routine time allocation (situational baseline) & B & Time\_Family, Time\_Personal, Time\_Sleep \\
2 & Time constraint / temporal response cost & B & Constraint\_Work, Constraint\_Household, Pressure\_Overwhelmed, Flex\_Work, Time\_Scarcity \\
3 & Threat appraisal & C & Aware\_Time, Helene\_Impact, Hurricane\_Experience \\
4 & Coping appraisal & C & Prep\_Stress, Prep\_Time \\
5 & Protective behavior & C & Prep\_Start, Prep\_Action\_Time \\
6 & Temporal impact & C & Personal\_Disruption, Time\_Spent\_Dependents \\
\bottomrule
\end{tabularx}
\end{table*}

\begin{figure}[t]
  \centering
  \includegraphics[width=\columnwidth]{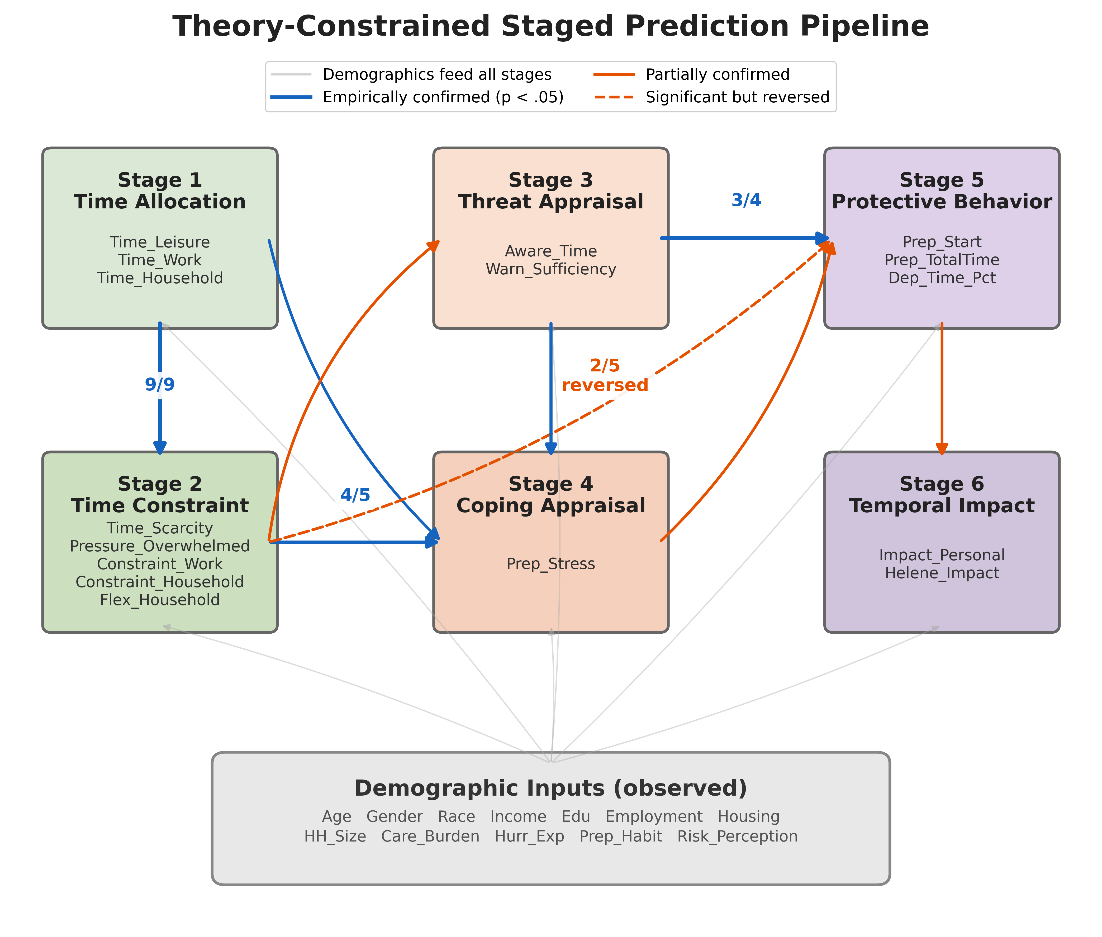}
  \caption{Theory-constrained staged prediction pipeline. The six-stage PMT cascade maps Block A demographic inputs through temporally ordered constructs to hurricane preparation outcomes. Edges retained in the PMT-constrained graph are shown; validated edges (Spearman sign-concordant) are highlighted.}
  \label{fig:pipeline}
\end{figure}

We instantiated the cascade as a PMT-constrained co-occurrence graph built from the 757 training records. Edges were retained only between source and target fields in which the source occupied an upstream PMT stage relative to the target. This procedure yielded 204 source nodes, 9,605 weighted edges, and 344 permitted source-to-target field pairs, with edge weights estimated as conditional cell counts. To assess whether the graph's directional assumptions held empirically, we tested each edge for agreement between its theoretical prediction and the observed Spearman correlation in the training data. This produced two operational variants: the full PMT-constrained graph, retaining every theoretically licensed edge with its empirical weight; and a validated-edge subgraph, restricted to edges whose theoretical sign matches the empirical Spearman correlation. The graph organizes evidence rather than enforcing a fixed inference path (Figure~\ref{fig:pipeline}): it specifies which conditional relationships are theoretically licensed, while leaving the integration of that evidence either to a staged process or to a single model call.

\section{LLM-Based Methods for Survey Research Operations}

A survey-research project unfolds through a sequence of five operational stages: designing the instrument, selecting the sample, pilot-testing the questionnaire, collecting and cleaning the data, and analyzing it. As introduced in Table~\ref{tab:framework}, we mapped a series of LLM operations to each of the five stages. For each stage we specify the LLM's input, output, and validation strategy, and whether the operation is evaluated as a controlled experiment (Stages 3 and 4) or as a demonstration grounded in real survey data (Stages 1, 2, and 5).

Methods for predicting survey responses from a respondent's persona, defined here as the Block A demographic profile, vary along two design choices: what task-specific evidence the language model receives at inference time, and how that evidence is structured. We evaluate seven methods spanning this design space; their prompt configurations are summarized in Table~\ref{tab:methods}.

Two baseline methods give the model no task-specific evidence and rely on its pretrained knowledge alone. The \textbf{Zero-Shot LLM (ZS-LLM)} issues the prediction request together with the persona and the survey question. The \textbf{Few-Shot LLM (FS-LLM)} extends the same prompt by inserting five complete training respondents drawn at random as illustrative input-output pairs.

Three retrieval-augmented methods follow the RAG pattern: at inference time an external information store is queried for evidence relevant to the target persona, and the retrieved evidence is inserted into the prompt before the model is asked to answer \citep{lewis2020}. The three methods share this overall pattern and differ in the structure of the evidence store, providing a controlled comparison of three progressively more informative organizations of the same training data. The \textbf{Embedding-Retrieval LLM (ER-LLM)} is the standard RAG configuration: the store is a flat index of the 757 training-set respondents, each represented as a numerical embedding; retrieval selects the training respondents whose embeddings lie nearest the target's, treating the training set as an unstructured collection. The \textbf{Graph-Retrieval LLM (GR-LLM)} replaces the flat index with a knowledge graph in which respondent attributes and survey answers are nodes connected by weighted edges recording empirically observed co-occurrences, so that retrieval traces a relational structure rather than a similarity score alone. The \textbf{Staged Theory-Informed LLM (Staged-TLM)} operates on the PMT-constrained knowledge graph, with retrieval proceeding one PMT stage at a time and each stage's point estimate becoming an input to the next. The three methods together isolate three structural increments: the addition of semantic similarity (ER-LLM), of empirical relational structure (GR-LLM), and of theoretical direction (Staged-TLM).

The remaining two methods hold the PMT-constrained graph as the evidence source and vary how the model integrates retrieved evidence. The \textbf{Marginal Theory-Informed LLM (Marginal-TLM)} supplies the model with the graph's full set of marginal and conditional probability distributions in a single prompt, so that the model weighs the evidence in one integration step rather than committing to staged point estimates. The \textbf{Anchored Marginal Theory-Informed LLM (A-TLM)}, the configuration we propose, extends Marginal-TLM with two persona-targeted signals: the five training respondents nearest the target persona on key demographics (peer examples), and, when the target persona meets at least two vulnerability criteria, the empirical mean difference in each outcome between the compound-vulnerable training subsample and the full training sample, conveyed as a soft directional cue rather than a numerical override.

Stage 3 evaluates all seven LLM methods. The Stage 4 main-text comparison evaluates five of them (ZS-LLM, FS-LLM, Staged-TLM, Marginal-TLM, A-TLM) against three classical imputation baselines: inverse-probability-weighted multiple imputation (IPW/MI; \citealt{seaman2012}), multivariate imputation by chained equations with predictive mean matching (MICE+PMM; \citealt{vanbuuren2011}), and a random-forest iterative imputer (missForest; \citealt{stekhoven2012}). The remaining two LLM methods (ER-LLM, GR-LLM) and a Validated-Edge Theory-Informed LLM (VE-TLM) are reported in supplementary materials. All LLM calls used \texttt{claude-sonnet-4.5} at temperature 0.1.

\begin{table*}[!ht]
\centering
\caption{Learning Contexts in LLMs: Method Descriptions and Illustrative Examples.}
\label{tab:methods}
\footnotesize
\setlength{\tabcolsep}{4pt}
\begin{tabularx}{\textwidth}{L{1.8cm} X L{3.0cm}}
\toprule
\textbf{Method} & \textbf{Description} & \textbf{Example} \\
\midrule
ZS-LLM & Uses only pretrained knowledge and the task instruction; no task-specific examples in the prompt. & Predicts 16 Block-B/C answers from Block A demographics alone; no training-set evidence supplied. \\
FS-LLM & Augments the zero-shot prompt with five randomly selected complete training respondents as input--output pairs. & Five training respondents inserted verbatim before the prediction request. \\
ER-LLM & Queries a dense-vector flat index at inference time; top-$k$ semantically similar training narratives appended to prompt (vanilla RAG). & LightRAG encoder retrieves nearest training respondents by embedding similarity. \\
GR-LLM & Queries a data-driven co-occurrence graph; conditional cell distributions consistent with the target persona returned as evidence (graph-RAG). & Weighted graph queried for cell distributions consistent with target demographics; cells summarized in prompt. \\
Staged-TLM & Traverses the PMT-constrained graph in a fixed six-stage causal cascade; each stage commits to a point estimate that conditions the next. & Stage 1 predicts time use; Stage 2 predicts constraints conditional on Stage 1; cascade continues through coping appraisal, protective behavior, and temporal impact. \\
Marginal-TLM & Summarizes the full PMT-constrained graph as marginal/conditional distributions; all evidence supplied in a single prompt. & Prompt reports distribution of training respondent answers by demographic profile and upstream PMT-stage values; model returns all 16 predictions in one call. \\
\textbf{A-TLM} & Extends Marginal-TLM with (i) the $k=5$ nearest training peers and (ii) empirical subgroup-shift parameters for compound-vulnerable personas as a soft directional signal. & Peer examples appended; for compound-vulnerable targets, per-item mean difference between vulnerable and full training samples also included in prompt. \\
\bottomrule
\end{tabularx}
\end{table*}

\subsection{Stage 1: Questionnaire Design via the LLM Instrument Audit}
Before fielding a survey, the research team must verify that the instrument measures the theoretical constructs of interest, a step traditionally performed by manual expert review. This stage automates that audit with a single LLM call. We provided the LLM with the 16 outcome items together with the full specification of the nine PMT constructs, namely Perceived Severity, Perceived Vulnerability, Fear Arousal, Prior Experience, Response Efficacy, Self-Efficacy, Response Cost, Protection Motivation, and Protective Behavior, and required a structured response covering per-construct adequacy scores (1 to 5), identified construct gaps with literature-grounded recommendations for additional items, identified redundancies, and an overall verdict on instrument sufficiency.

Validation proceeded along three axes. First, we assessed the face validity of identified gaps by comparing them against published PMT-based hurricane instruments \citep{floyd2000,lazo2015,slovic2007}. Second, each flagged redundancy was independently inspected in the codebook. Third, we assessed whether the audit's verdict correctly reconciled the full PMT framework with the instrument's actual research focus on time-constrained disaster preparation \citep{wang2026}.

\subsection{Stage 2: Sample Selection via the LLM Sample-Coverage Prior}
Survey samples almost never match the population on every dimension, and the resulting under-representation distorts inferences. This stage uses the LLM as a pre-fielding planning tool: it forecasts which subgroups are likely to be under-represented in a post-hurricane survey, and that forecast is then validated against the empirical gap measured after fielding.

The LLM was prompted to predict, from the published disaster-survey literature \citep{fussell2014}, which demographic subgroups are typically under-represented in post-hurricane convenience samples, without access to the Hurricane Milton sample. The model returned a ranked list of subgroups with literature-grounded justifications. We then computed the empirical gap as the Florida ACS population percentage minus the Hurricane Milton sample percentage for each of eight subgroups. Validation used the Spearman rank correlation between the LLM's prior ranking and the empirical gap ranking as the primary performance metric, with a threshold of $\rho > 0.5$ as evidence of a useful pre-fielding planning tool.

\subsection{Stage 3: Pilot Testing via Response Prediction}
Pilot testing has historically required recruiting real respondents to verify that items produce variation, check that planned cross-tabulations will populate at usable sample sizes, refine ambiguous wording, and budget statistical power before committing to full fielding. We tested whether an LLM can fill the same design-validation role by generating simulated responses conditioned only on a respondent's demographic profile. The purpose is design support, not data fabrication: simulated responses inform questionnaire and analysis-plan decisions before fielding and are not introduced into the final analytic dataset.

Each of the seven LLM methods was asked to predict the full set of 16 Block B and Block C answers for each of the 189 held-out validation respondents, conditioning only on Block A demographics. The full A-TLM specification, including the 0.10 Likert-level inclusion threshold for the subgroup-shift signal, is given in the supplementary material. Evaluation was restricted to ordinal-mappable cells; fields containing skip-logic missing values were excluded. For each method we report mean absolute error (MAE) on the ordinal Likert mapping, within-1 accuracy, quadratic-weighted Cohen's $\kappa$, and semantic similarity computed via the SentenceTransformer all-MiniLM-L6-v2 model.

\subsection{Stage 4: Data Collection via Simulated-Missingness Imputation}
Real surveys rarely collect complete responses. Missingness is typically related both to the answer itself and to respondent characteristics, a pattern known as missing not at random (MNAR). This stage tests whether an LLM guided by theory and demographic anchoring can impute missing responses more reliably than classical methods under progressively demanding conditions.

We evaluated each method under four reproducible missingness mechanisms applied to the 189-respondent validation set (seed$=42$). \textbf{Mechanism S1} (MCAR; sanity baseline) deleted 20\% of Block B and Block C cells uniformly at random. \textbf{Mechanism S2} (MAR; demographic-driven) assigned higher deletion probability to low-income and minority respondents, conditional on Block A demographics. \textbf{Mechanism S3} (MNAR-moderate; response-driven) elevated Block C deletion probability for respondents with lower preparation scores. \textbf{Mechanism S4} (MNAR-severe; block-wise) deleted the entire Block C for respondents meeting at least two vulnerability criteria, simulating permanent displacement; this produced 60\% Block C missingness concentrated among compound-vulnerable respondents.

IPW/MI was implemented following \citet{seaman2012}: respondents with at least 50\% block-level missingness were excluded and inverse-probability weighted using logistic-regression weights conditional on Block A; included respondents with isolated missing items received $M=5$ multiple imputations with the inverse probability of inclusion as a covariate; Rubin's Rules combined results and a sandwich variance estimator was used for the weighted-mean estimator. A component ablation evaluated four A-TLM variants on Block C across the four scenarios: Marginal-TLM alone, plus peer examples only, plus vulnerability deltas only, and the full A-TLM combining both signals.

For each method, scenario, and block, the primary metrics are root-mean-square error (RMSE) and signed bias on deleted cells. Secondary metrics include 95\% confidence-interval coverage via Rubin pooling, symmetric Kullback-Leibler divergence between imputed and ground-truth Likert distributions, and the $\ell_2$ norm of the coefficient difference between OLS fits on imputed versus ground-truth datasets. All metrics are computed on both the full sample and the compound-vulnerable subgroup. Five preregistered sanity checks gated the results: (1) positive between-imputation variance under Rubin pooling; (2) IPW weights bounded at 10; (3) zero out-of-range imputed values on the 1 to 5 scale; (4) missingness patterns in MNAR scenarios significantly different from MCAR and MAR by Fisher's exact test; and (5) exact preservation of ground-truth values wherever the deletion mask was zero.

\subsection{Stage 5: Data Analysis via the Graph-Grounded Survey Assistant}
Survey datasets are typically delivered as static files that require statistical expertise to query. This stage tests whether routing user questions through an LLM that retrieves cells of a knowledge graph can let non-specialist stakeholders receive numerically grounded answers. We deployed a web chatbot that mediates between natural-language survey questions and the PMT-constrained co-occurrence graph built from the 757 training records. The pipeline parses each question via an LLM call to identify relevant survey variables, retrieves the corresponding marginal distributions, cross-tabulations, and conditional answer patterns from the graph, and returns a data-grounded answer that explicitly references sample sizes, percentages, and PMT-stage interpretations.

We validated the system against three criteria: every numeric claim traceable to a cell of retrieved graph evidence (quantitative grounding); findings linked to PMT-stage interpretations where applicable (theory integration); and the assistant declines to answer when retrieved evidence is insufficient rather than confabulating a response (epistemic honesty).

\FloatBarrier\section{Experimental Results}

\subsection{Stage 1: Construct-Adequacy Audit}
We asked the LLM to score the coverage of each of the nine PMT constructs by the 16-item Hurricane Milton instrument on a 1 to 5 adequacy scale and to recommend additional items for each construct scoring below the threshold. Per-construct scores are reported in Table~\ref{tab:audit}. These results should be interpreted as an illustrative demonstration of LLM-assisted audit capability rather than a definitive validation, given that no formal inter-rater reliability comparison was conducted against an expert-panel baseline.

\begin{table*}[!ht]
\centering
\caption{LLM Adequacy Audit of Hurricane Milton Instrument Against Nine PMT Constructs.}
\label{tab:audit}
\small
\begin{tabularx}{\textwidth}{L{3.2cm} C{1.0cm} X}
\toprule
\textbf{PMT Construct} & \textbf{Score (1--5)} & \textbf{Instrument Items Mapped} \\
\midrule
Perceived Severity       & 1 & None \\
Perceived Vulnerability  & 2 & Helene\_Impact \\
Fear Arousal             & 2 & Prep\_Stress \\
Prior Experience         & 2 & Helene\_Impact \\
Response Efficacy        & 1 & None \\
Self-Efficacy            & 2 & Flex\_Household \\
Response Cost (temporal) & \textbf{4} & Prep\_TotalTime, Dep\_Time\_Pct, Impact\_Personal, Constraint\_Work, Constraint\_Household, Time\_Scarcity, Pressure\_Overwhelmed \\
Protection Motivation    & 1 & None \\
Protective Behavior      & 3 & Aware\_Time, Prep\_Start, Prep\_TotalTime \\
\bottomrule
\end{tabularx}
\end{table*}

Only one construct, the temporal dimension of response cost, received a score of 4 or higher; it was mapped to seven items spanning routine time allocation and time constraint. Three constructs (Perceived Severity, Response Efficacy, and Protection Motivation) scored at the floor of the scale because the instrument contains no items mapped to them. The audit's overall verdict was that the instrument is fit for the project's research focus on time-constrained disaster preparation but does not support full PMT pathway testing, an assessment consistent with the instrument's original design goals \citep{wang2026}. Eight literature-grounded gap-item recommendations accompanied the per-construct scores. The same assessment by an expert reviewer would typically require several hours of manual cross-referencing; the LLM returned an equivalent scoping diagnostic in a single inference call.

\subsection{Stage 2: Sample-Coverage Prior}
We asked the LLM to rank demographic subgroups most likely to be under-represented in a post-hurricane convenience sample, conditioning only on the published disaster-survey literature and not on the Hurricane Milton respondents. We then computed the actual deviation between the Milton sample composition and the Florida adult population from the ACS 5-year estimates. Both quantities are reported in Table~\ref{tab:sample}.

\begin{table*}[!ht]
\centering
\caption{Sample Composition versus Florida ACS Reference. Positive gaps indicate under-representation in the Milton sample relative to the population. Rank 1 = most under-represented; Rank 8 = most over-represented.}
\label{tab:sample}
\small
\begin{tabular}{lrrrrr}
\toprule
\textbf{Subgroup} & \textbf{Milton (\%)} & \textbf{ACS Florida (\%)} & \textbf{Gap (ACS$-$Milton, pp)} & \textbf{LLM Rank} & \textbf{Empirical Rank} \\
\midrule
White non-Hispanic & 66.7 & 52.0 & $-14.7$ & 7 & 7 \\
Hispanic           & 10.8 & 26.0 & $+15.2$ & 2 & 1 \\
Black              & 17.0 & 16.0 & $-1.0$  & 1 & 4 \\
Asian              &  3.5 &  3.0 & $-0.5$  & 6 & 3 \\
Income $<\$35$k    & 27.6 & 24.0 & $-3.6$  & 3 & 6 \\
Income $>\$150$k   & 10.5 & 15.0 & $+4.5$  & 8 & 2 \\
Renter             & 34.1 & 33.0 & $-1.1$  & 4 & 5 \\
Disabled (18+)     & 29.1 & 14.0 & $-15.1$ & 5 & 8 \\
\midrule
\multicolumn{6}{l}{\textit{Note.} Spearman $\rho$ between LLM prior ranking and empirical gap ranking $= 0.12$.}\\
\bottomrule
\end{tabular}
\end{table*}

The Spearman rank correlation between the LLM's literature-prior ranking and the empirical gap ranking was $\rho = 0.12$, indicating limited correspondence between what the literature predicts and what the online-panel recruitment produced. The LLM correctly identified Hispanic respondents as among the most under-represented, consistent with the disaster-survey literature \citep{fussell2014}, but ranked Black respondents as the most under-represented when their sample share was within one percentage point of the population reference. The two largest discrepancies involved over-representation of respondents reporting any disability (29.1\% versus 14.0\%) and under-representation of high-income respondents (10.5\% versus 15.0\%); neither pattern is anticipated by the disaster-survey literature, and both are consistent with online-panel self-selection dynamics rather than disaster-specific coverage gaps. The value of the LLM prior in this stage lies less in the accuracy of its rankings and more in establishing a structured baseline: the divergences between the prior and the observed sample composition identify subgroups that warrant targeted recruitment attention in subsequent fielding waves and that would not have been flagged by the literature consensus alone.

\subsection{Stage 3: Response Prediction for Pilot Testing}
Each of the seven LLM methods was asked to predict the full set of 16 Block B and Block C answers for each of the 189 held-out validation respondents, conditioning only on Block A demographic profiles. Mean absolute error on the ordinal coding, computed separately for Block B and Block C, is reported in Table~\ref{tab:mae}. These results reflect the performance of each method as a design-support tool for pilot testing, not as a substitute for primary data collection.

\begin{table}[!ht]
\centering
\caption{MAE on Response Prediction, 189 Validation Respondents (lower is better). Bold = best.}
\label{tab:mae}
\small
\begin{tabular}{lrrr}
\toprule
\textbf{Method} & \textbf{Blk B} & \textbf{Blk C} & \textbf{Combined} \\
\midrule
\textbf{Marginal-TLM} & \textbf{0.949} & \textbf{1.067} & \textbf{0.993} \\
A-TLM      & 0.960 & 1.074 & 1.003 \\
GR-LLM     & 0.952 & 1.131 & 1.016 \\
Staged-TLM & 0.953 & 1.156 & 1.030 \\
FS-LLM     & 0.984 & 1.117 & 1.031 \\
ZS-LLM     & 0.999 & 1.159 & 1.060 \\
ER-LLM     & 1.081 & 1.123 & 1.097 \\
\midrule
\multicolumn{4}{l}{\textit{Note.} Cell-weighted avg; skip-logic cells excluded.}\\
\bottomrule
\end{tabular}
\end{table}

The two blocks differed substantially in predictability. Block B MAE was lower than Block C MAE across all seven methods because routine time use is largely a function of demographic role characteristics, such as age, household size, employment type, and schedule structure, that are directly recoverable from Block A. Hurricane-preparation answers depend additionally on perceived threat, prior experience, household resources, and local risk perception, none of which are recoverable from demographics alone. Block C therefore shows both higher overall error and a wider spread between the best (Marginal-TLM, 1.067) and worst (ZS-LLM, 1.159) methods.

The seven methods represent a sequence of design choices, each adding a different layer of evidence. ER-LLM, the standard retrieval-augmented configuration, was the worst-performing method (combined MAE 1.097), suggesting that nearest-neighbor retrieval without relational or theoretical structure can mislead the model when respondents with similar narratives differ in the demographic dimensions that drive hurricane behavior. GR-LLM recovers the accuracy lost by ER-LLM by replacing the embedding index with a co-occurrence graph (1.016). Staged-TLM layers a directional PMT cascade on the same graph but traverses it one stage at a time; the resulting error propagation causes it to underperform GR-LLM (1.030). Marginal-TLM integrates the full PMT-organized evidence in a single call, achieving the lowest combined MAE (0.993). A-TLM performs on par with Marginal-TLM (1.003): the persona-targeted signals it adds do not further reduce prediction error at this stage, where all respondents are observed, but become consequential at Stage~4 where compound-vulnerable respondents are systematically absent. Across all methods, the absolute MAE values of approximately one ordinal level on a five-point scale are appropriate for design-support decisions such as prototyping cross-tabulations and identifying low-frequency cells, but fall short of what would be required for population-level inferential precision.

\subsection{Stage 4: Imputation Under Simulated Missingness}

\subsubsection{Main Comparison}
\begin{figure}[t]
  \centering
  \includegraphics[width=\columnwidth]{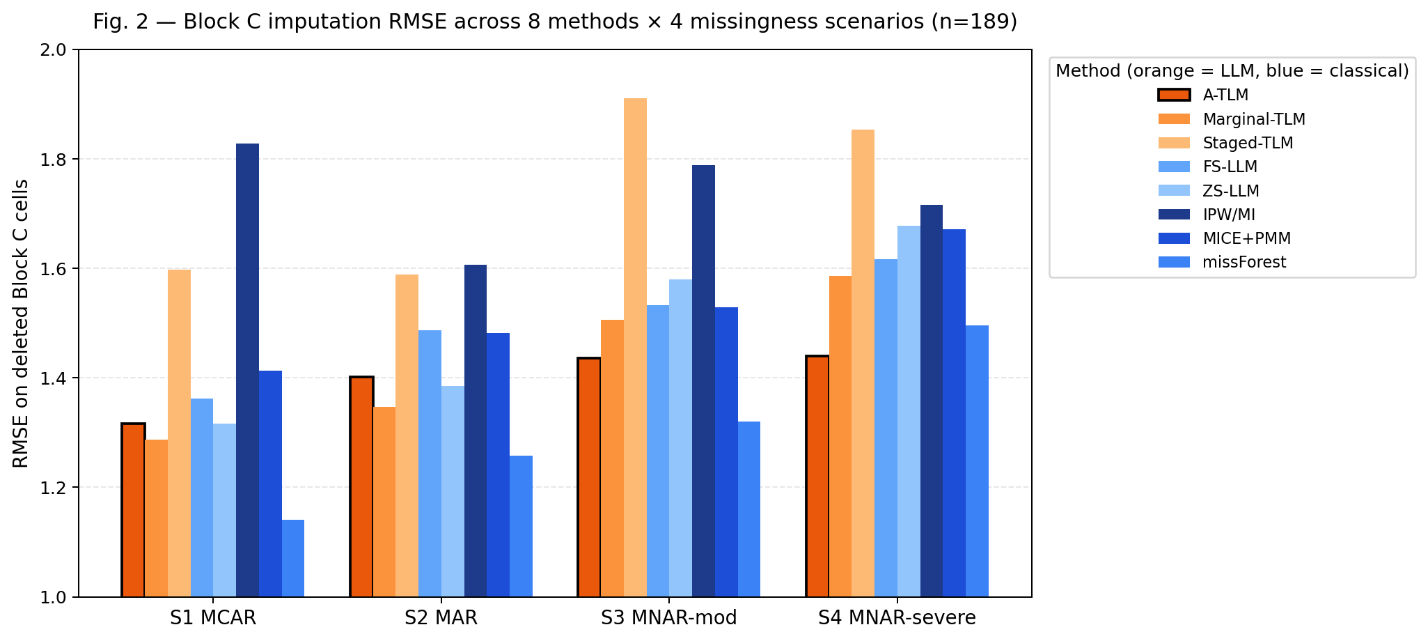}
  \caption{Block C imputation RMSE across four missingness mechanisms. A-TLM achieves the lowest RMSE under S4, the scenario most representative of post-disaster block-wise displacement.}
  \label{fig:rmse}
\end{figure}

Root-mean-square error and signed bias on the deleted Block C cells across the four missingness mechanisms are reported in Table~\ref{tab:imputation} and visualized in Figure~\ref{fig:rmse}. These results are directionally informative; the performance differences between closely ranked methods, particularly under S4, should be interpreted with the caveat that the compound-vulnerable validation subsample comprises 72 respondents and formal inferential statistics on the RMSE differences were not computed.

\begin{table*}[!ht]
\centering
\caption{Block C Imputation: RMSE and Signed Bias Across the Four Missingness Mechanisms. Bold denotes the lowest RMSE under S4.}
\label{tab:imputation}
\scriptsize
\setlength{\tabcolsep}{4.5pt}
\begin{tabular}{lrrrrrrrr}
\toprule
\textbf{Method} & \textbf{S1 RMSE} & \textbf{S1 Bias} & \textbf{S2 RMSE} & \textbf{S2 Bias} & \textbf{S3 RMSE} & \textbf{S3 Bias} & \textbf{S4 RMSE} & \textbf{S4 Bias} \\
\midrule
ZS-LLM       & 1.316 & $+$0.526 & 1.385 & $+$0.430 & 1.580 & $+$0.866 & 1.677 & $+$0.611 \\
FS-LLM       & 1.361 & $+$0.164 & 1.486 & $+$0.000 & 1.533 & $+$0.536 & 1.616 & $+$0.096 \\
Staged-TLM   & 1.597 & $+$0.603 & 1.588 & $+$0.616 & 1.910 & $+$1.031 & 1.853 & $+$0.768 \\
Marginal-TLM & 1.287 & $+$0.328 & 1.347 & $+$0.302 & 1.506 & $+$0.742 & 1.586 & $+$0.485 \\
A-TLM        & 1.316 & $+$0.026 & 1.402 & $-$0.081 & 1.436 & $+$0.392 & \textbf{1.439} & $-$0.121 \\
IPW/MI       & 1.828 & $+$0.283 & 1.607 & $-$0.177 & 1.788 & $+$0.511 & 1.716 & $-$0.096 \\
MICE+PMM     & 1.412 & $+$0.071 & 1.482 & $+$0.012 & 1.529 & $+$0.330 & 1.671 & $+$0.145 \\
missForest   & 1.141 & $-$0.043 & 1.258 & $-$0.140 & 1.320 & $+$0.216 & 1.496 & $-$0.631 \\
\bottomrule
\end{tabular}
\end{table*}

\begin{figure}[t]
  \centering
  \includegraphics[width=\columnwidth]{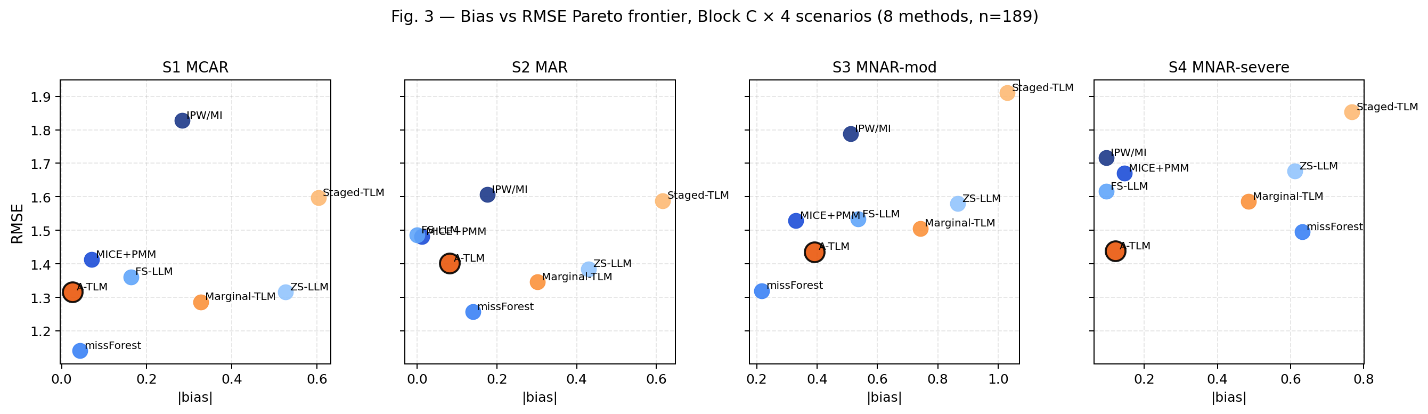}
  \caption{Bias-RMSE frontier on Block C imputation by missingness scenario. Methods below and to the left of A-TLM are preferable on both metrics simultaneously; no method clearly dominates A-TLM in any scenario.}
  \label{fig:frontier}
\end{figure}

A-TLM achieved a lower RMSE than IPW/MI across all four scenarios, a lower RMSE than MICE+PMM across all four scenarios, and the lowest RMSE of any method evaluated in S4 (1.439, compared with 1.496 for missForest). missForest achieved the lowest RMSE in S1, S2, and S3 but the largest absolute bias of any method in S4 ($-0.631$ ordinal levels), reflecting a systematic under-prediction of compound-vulnerable respondents that would distort any policy-relevant marginal-mean estimate derived from the imputed data. Staged-TLM had the highest RMSE in three of the four scenarios. The bias-versus-RMSE frontier in Figure~\ref{fig:frontier} shows that no method clearly dominates A-TLM on both metrics simultaneously in any scenario, though the margin in S4 is modest and future evaluations with larger validation samples will be needed to confirm this ordering.

\subsubsection{Component Ablation}
To attribute A-TLM's improvement over Marginal-TLM to its two added signals, we ran four ablations on Block C imputation: Marginal-TLM with neither signal, with peer examples only, with the vulnerability cue only, and with both. Signed bias on deleted Block C cells is reported in Table~\ref{tab:ablation}.

\begin{table*}[!ht]
\centering
\caption{Component Ablation: Signed Bias on Block C Across the Four Missingness Scenarios.}
\label{tab:ablation}
\small
\begin{tabular}{lrrrr}
\toprule
\textbf{Variant} & \textbf{S1} & \textbf{S2} & \textbf{S3} & \textbf{S4} \\
\midrule
Marginal-TLM (baseline) & $+$0.353 & $+$0.337 & $+$0.794 & $+$0.500 \\
$+$ Peer examples only  & $+$0.138 & $+$0.105 & $+$0.567 & $+$0.379 \\
$+$ Vulnerability cue only & $+$0.276 & $+$0.233 & $+$0.691 & $+$0.409 \\
$+$ Both signals (A-TLM)   & $+$0.164 & $+$0.105 & $+$0.588 & $+$0.298 \\
\bottomrule
\end{tabular}
\end{table*}

Adding peer examples to Marginal-TLM reduced absolute bias by 0.22 ordinal levels in S1, 0.23 in S2, 0.23 in S3, and 0.12 in S4. Adding the vulnerability cue alone produced smaller but consistent reductions across all four scenarios. The two signals combined super-additively only in S4 (A-TLM bias 0.298 versus peer-only bias 0.379), the scenario in which compound-vulnerable respondents are entirely absent from the observed evidence. There, the vulnerability cue carries non-redundant information that peer examples alone cannot supply.

\subsubsection{Subgroup-Stratified Bias}
\begin{figure}[t]
  \centering
  \includegraphics[width=\columnwidth]{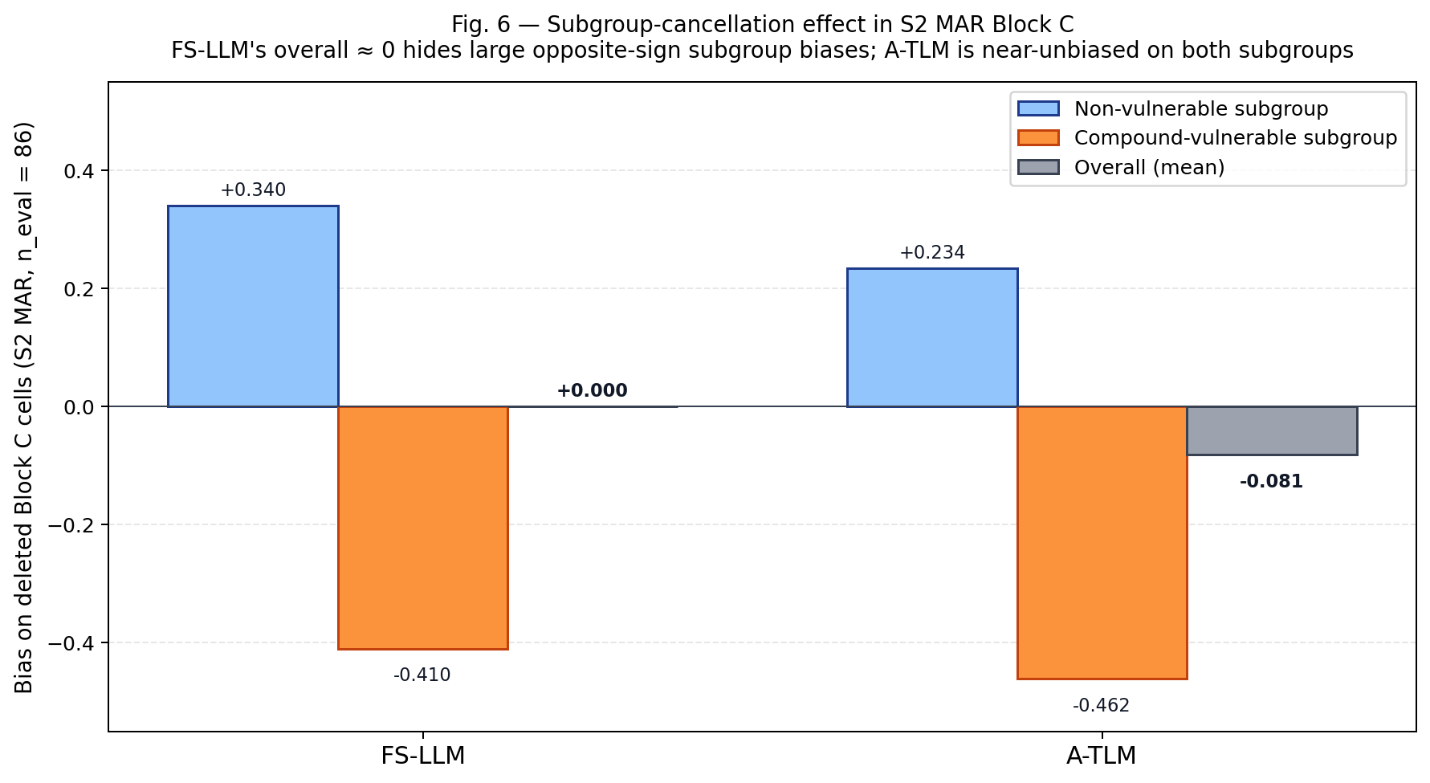}
  \caption{Subgroup-stratified bias on Block C imputation under S2 (MAR). Bars show signed bias separately for compound-vulnerable ($n=39$) and non-vulnerable ($n=47$) validation respondents. Near-zero overall bias in several methods conceals opposing errors of opposite sign across subgroups.}
  \label{fig:subgroup}
\end{figure}

A near-zero overall bias can mask substantial subgroup-specific errors of opposite sign. We computed bias separately for non-vulnerable and compound-vulnerable respondents under S2 on the deleted Block C cells (Figure~\ref{fig:subgroup}).

Under S2, FS-LLM produced an overall bias of essentially zero on Block C. This figure was the cancellation of a $+0.340$-level over-prediction among non-vulnerable respondents ($n=47$) and a $-0.410$-level under-prediction among compound-vulnerable respondents ($n=39$). A-TLM produced a smaller but qualitatively similar pattern ($+0.234$ and $-0.462$ respectively). The cancellation is a general feature of LLM imputation under this scenario rather than a property of any single method. For policy-relevant applications where estimates for high-missingness subgroups are the inferential target, reporting overall bias alone is insufficient; subgroup-stratified estimates should be reported alongside aggregate figures as standard practice.

\subsubsection{Routine Time-Use Items (Block B)}
All theory-informed methods outperformed the baseline LLM methods on Block B. However, A-TLM offered no measurable improvement over Marginal-TLM, with absolute-bias differences below 0.05 ordinal levels across all four scenarios. The persona-targeted signals in A-TLM contribute specifically to Block C, where respondent-level heterogeneity not captured by demographics is the binding constraint. Routine time-use behavior is sufficiently predicted by Block A demographic role characteristics that the model's pretraining prior already provides adequate signal. Detailed Block B results are reported in supplementary Table S3.

\subsection{Stage 5: Knowledge-Graph Chatbot}
We deployed the Graph-Grounded Survey Assistant, a web chatbot that retrieves evidence from the PMT-constrained co-occurrence graph in response to natural-language questions and constructs an answer grounded in the retrieved cells. We evaluated three captured exchanges against the three pre-specified criteria.

In one exchange, the user asked whether respondents with flexible work schedules report less preparation stress. The assistant retrieved the relevant cross-tabulation and reported, with sample sizes, that 26.2\% of respondents with ``Very flexible'' schedules also reported ``Not at all stressed,'' compared with 9.6\% among respondents with ``Not flexible'' schedules. Each percentage was traceable to a specific retrieved cell, and the assistant contextualized the finding within PMT by noting that schedule flexibility (a Stage 2 time-constraint item) moderates coping appraisal. This exchange satisfied both the quantitative-grounding and theory-integration criteria.

\begin{figure}[t]
  \centering
  \includegraphics[width=\columnwidth]{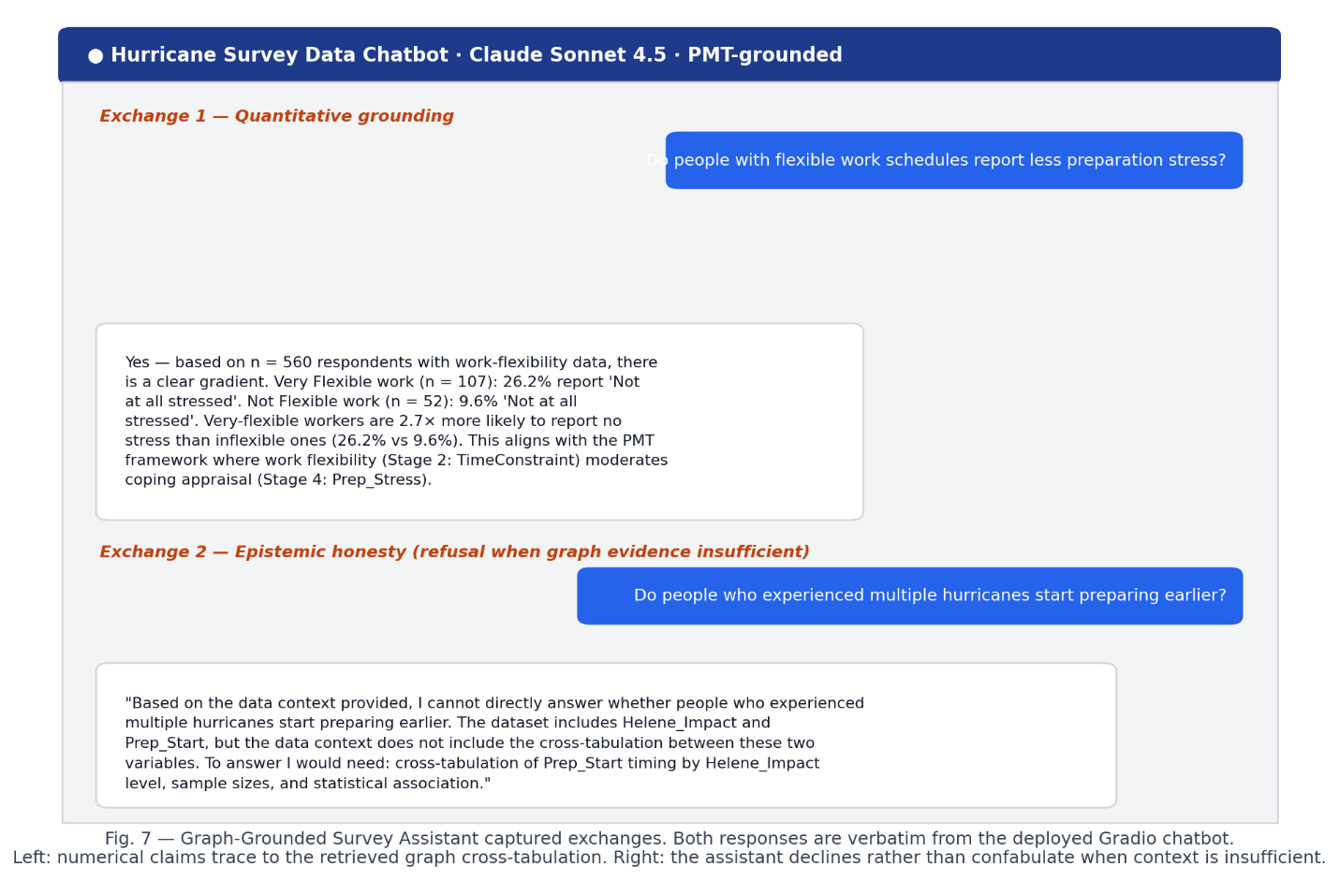}
  \caption{Captured exchanges from the Graph-Grounded Survey Assistant. Top: successful quantitative retrieval with PMT-stage contextualization. Bottom: epistemic refusal when retrieved evidence is insufficient.}
  \label{fig:chatbot}
\end{figure}

In a second exchange, the user asked whether respondents with prior multi-hurricane experience start preparing earlier. The assistant retrieved the marginal distributions of the two relevant items but found no joint cross-tabulation in the retrieved evidence and declined to answer (Figure~\ref{fig:chatbot}). This refusal behavior is architectural: the system is constrained to ground every numeric claim in evidence present in the prompt, and in its absence states which additional data would be required. This exchange satisfied the epistemic-honesty criterion.

\FloatBarrier\section{Discussion}

\subsection{Contributions to Knowledge}

This paper operationalizes a five-stage framework for embedding large language models within the survey-research workflow and evaluates each stage on the 2024 Hurricane Milton preparedness survey of Florida residents \citep{wang2026}. The contribution claimed here is the operationalization itself: each of the five stages carries a reproducible script grounded in the same dataset and the same PMT-constrained co-occurrence graph, and every numerical claim traces to a specific table or figure. We discuss the contributions at three levels: the conceptual framework, the methodological innovations, and the broader implications for how AI systems should complement human expertise in survey science.

\subsubsection{A Unified Operational Framework for LLM-Augmented Survey Research}
Prior work on LLMs in survey contexts has largely addressed individual operations in isolation: whether silicon sampling can replicate human response distributions \citep{argyle2023,sun2024a}, whether LLM-assisted interviewing improves open-ended data quality \citep{barari2025,kaiyrbekov2025}, or whether AI-generated responses introduce systematic error \citep{bisbee2024,ashwin2025}. No prior study evaluates LLM contributions across the full survey lifecycle on a single shared empirical testbed. By anchoring all five stages to the same dataset and theoretical structure, this framework enables cumulative evaluation: subsequent studies can extend or challenge individual stages without rebuilding the evaluation infrastructure from scratch.

The framework also reframes a normative question that has run through this literature. Rather than asking whether LLMs can replace human judgment in survey research, it asks where in the workflow LLMs can serve as a reliable assistant, under what conditions, and subject to what constraints. This reframing aligns with the position that AI systems are most productively understood as tools that augment human capacity in well-defined tasks rather than as general substitutes for domain expertise \citep{crockett2025,sarstedt2024}. The five-stage structure makes that position operationally concrete by assigning a human oversight role at each stage alongside the LLM operation.

\subsubsection{Theory-Constrained Retrieval as a Methodological Contribution}
The experimental results across Stages 3 and 4 establish that how the evidence store is organized matters as much as the decision to retrieve at all. Standard RAG grounds model outputs in empirical evidence but treats that evidence as an unstructured collection. Replacing the flat embedding index with a co-occurrence graph that encodes relational structure recovers accuracy lost by embedding-based retrieval. Crucially, the gain from theory-constrained graph retrieval is realized only when evidence is integrated in a single model call. Staged-TLM, which applies the same PMT structure but commits to a point estimate at each step, amplifies bias across the cascade rather than correcting it, confirming that the theoretical structure is not itself the source of gain unless the model can weigh all evidence simultaneously. This finding connects to broader observations in NLP on the risks of committing to intermediate outputs before all evidence is integrated \citep{huang2024}, extending that insight to a structured social-science application.

In the survey methodology literature, the finding that theoretical structure improves imputation performance by shaping how auxiliary information is weighted is novel. PMT has served primarily as a post-hoc explanatory framework for preparedness behavior \citep{floyd2000,lazo2015,wang2026}, but it has not previously been operationalized as a retrieval constraint during missing-data imputation. The demonstration that this operationalization reduces signed bias on compound-vulnerable respondents represents a meaningful integration of substantive domain theory and computational method.

\subsubsection{Imputation Performance Under Disaster-Relevant Missingness}
The Stage 4 results surface a pattern with practical consequences for disaster research. Classical methods such as missForest achieve competitive overall RMSE but produce the largest absolute bias on compound-vulnerable respondents under block-wise displacement conditions. This occurs because aggregate RMSE can remain favorable even when errors of opposite sign cancel across respondent groups. A-TLM reduces this subgroup-differential bias by anchoring predictions to demographically matched peer examples and, when the target respondent meets multiple vulnerability criteria, to an empirical directional signal derived from the training distribution of respondents with comparable characteristics. Subgroup-stratified bias reporting is therefore proposed not as an optional sensitivity check but as a standard reporting practice for any imputation workflow applied to disaster survey data where high-missingness subgroups are also the primary inferential targets.

\subsubsection{LLMs as Bounded Research Assistants: Scope and Epistemic Honesty}
A recurring challenge in applied AI research is communicating what a system can and cannot do without either overstating its performance or dismissing genuine utility. This study addresses that challenge operationally by reporting, for each stage, both what the LLM contribution achieves and where it falls short. The Stage 2 result is illustrative: a literature-derived prior for sample coverage produced a Spearman correlation of 0.12 with the observed ACS gap. The informational content lay in the divergences: the two subgroups where the prior most differed from observation identified recruitment-mode artifacts that the disaster-survey literature does not describe and that would not have been flagged by the literature alone. The LLM is useful here not because its rankings were accurate but because its structured hypothesis provided a baseline that the empirical comparison could test and revise, consistent with \citealt{horton2023}'s positioning of LLMs as instruments for hypothesis generation and with \citealt{sarstedt2024}'s recommendation that silicon samples be used at upstream design stages.

The Stage 5 chatbot extends this principle architecturally. By constraining the system to ground every numerical claim in explicitly retrieved graph cells and to decline when those cells are absent, the chatbot converts hallucination from an inherent model tendency into a manageable design parameter. The refusal behavior demonstrated in Figure~\ref{fig:chatbot} is a feature, not a limitation: it makes the boundary between what the data support and what they do not visible to the user, directly addressing the epistemia risk identified by \citet{loru2025}.

\subsection{Limitations and Future Directions}

We address the principal limitations of this study alongside the research directions they motivate.

\subsubsection{Evaluation Scope and Generalizability}
The framework was evaluated on a single post-hurricane event in one U.S.\ state using an online panel. Performance in face-to-face field settings, other disaster types, or non-English survey environments is not yet established. LLM performance varies with the linguistic and cultural profile of the training corpus \citep{sun2024b}, and post-hurricane convenience samples in the United States are structurally distinct from samples collected through other modes or in other regions \citep{king2002,hao2022}. Within these bounds, the Hurricane Milton data provide an ecologically valid testbed grounded in a real post-disaster event with documented composition and ACS population benchmarks \citep{wang2026}. The single-event scope motivates a replication agenda: applying the same five-stage operations to post-disaster datasets from other events and settings will enable systematic evaluation of what generalizes and what is context-specific. Extending the framework to non-English contexts will require multilingual foundation models and evaluation datasets that reflect the linguistic diversity of global disaster-affected populations.

\subsubsection{Sample Scale and Statistical Precision}
The Stage 4 imputation benchmark for compound-vulnerable respondents rests on 72 validation respondents under the most demanding missingness scenario. The performance differences reported are directionally informative and consistent with a methodological pilot study, but formal inferential statistics such as bootstrap confidence intervals on RMSE differences were not computed. Future evaluations should target compound-vulnerable subsamples large enough to support such tests. Linking survey data to administrative records, such as disaster-relief registrant files, could also provide verified ground truth that goes beyond held-out validation-set simulation.

\subsubsection{Demonstration Versus Controlled Experiment}
Three of the five stages, the instrument audit, the sample-coverage prior, and the knowledge-graph chatbot, are evaluated as demonstrations grounded in real data rather than as controlled experiments with preregistered held-out ground truth. Their quantitative contributions are suggestive rather than definitive. Standardized evaluation protocols for these stages do not yet exist. For Stage 1, independent expert panels could audit the same instrument and provide a basis for measuring agreement with the LLM's recommendations. For Stage 5, a held-out question bank with adjudicated answers would enable precision and recall reporting for both correct retrieval and appropriate refusal behavior. Developing these protocols is a concrete priority for follow-on research.

\subsubsection{Synthetic Respondents: Appropriate Use and Limits}
The Stage 3 results show that the best-performing LLM configurations agree with held-out real respondents to within roughly one ordinal level on a five-point scale. This level of fidelity supports design-support tasks, including prototyping cross-tabulations, checking item variation, and calibrating power expectations, but falls short of what population-level inferential precision requires. Synthetic respondents are appropriately used in the design phase rather than the analytic phase. This distinction is especially valuable in rapid-onset disaster settings where access to target populations is constrained, but responsible deployment requires clear articulation of these limits and ongoing evaluation against real respondent data.

\subsubsection{Privacy and Deployment Infrastructure}
Routing survey microdata through commercial model APIs raises privacy considerations that vary by jurisdiction and institutional policy. This study uses anonymized experimental data and presents the framework as a research prototype. Translating it into operational deployment will require privacy-compliant infrastructure, audit-trail standards, and data-handling protocols that most survey organizations do not yet have in place. Locally deployed open-weight models offer a partial path forward. Beyond privacy, the computational resources and expertise required to implement LLM-augmented workflows may be out of reach for smaller research teams, particularly those working in resource-constrained settings. Developing practical guidelines, including IRB protocol templates and training resources for survey methodologists, is a collaborative task that deserves sustained institutional attention.

\FloatBarrier\section{Conclusion}

This paper demonstrates that large language models, embedded within a theoretically grounded and architecturally disciplined framework, can make meaningful and reproducible contributions at each stage of the survey-research process. The contributions are specific and bounded: they are strongest where the LLM operates within a structured evidence architecture that organizes what the model can retrieve and constrains what it can assert.

The central methodological finding is that how retrieval is organized determines whether the model assists or misleads. Grounding retrieval in a theory-derived causal structure and integrating all evidence in a single model call consistently outperforms unstructured similarity search and staged sequential reasoning. This result suggests that productive combination of domain theory and machine learning in social measurement depends on specifying precisely how theoretical structure shapes the evidence integration process, not merely on adding retrieval to a language model.

The finding that aggregate imputation metrics can mask systematic errors for specific respondent groups carries particular weight for disaster research, where the households most difficult to reach are often those most relevant to relief planning and recovery policy. Reporting imputation bias separately for groups defined by their missingness risk, rather than in aggregate alone, is a straightforward methodological standard whose adoption would improve the reliability of LLM-augmented survey analysis for policy-relevant applications.

Ultimately, the empirical responses of real survey participants remain the irreplaceable foundation of survey science. Every measurement, including those augmented by LLMs, is at best an estimate shaped by methodological choices, respondent behavior, and practical constraints. What LLMs can offer, when deployed responsibly within the kind of framework evaluated here, is help researchers make better use of the data they have: more efficiently at the design stage, with greater interpretive nuance under missing-data conditions, and more transparently at the analysis stage. Realizing that potential will require the same interdisciplinary rigor, ethical scrutiny, and commitment to human oversight that the best survey research has always demanded.

\FloatBarrier\section*{Declaration of Generative AI Use}
During the preparation of this manuscript, the authors made use of Claude Sonnet 4.6 (Anthropic) for assistance with language refinement and proofreading. The authors have reviewed and edited the output and take full responsibility for the content of this publication.

\section*{Acknowledgements}
This manuscript is based on work supported by the National Science Foundation under Grant No.\ 2505675 and No.\ 2440023. Any opinions, findings, and conclusions or recommendations expressed in this material are those of the authors and do not necessarily reflect the views of the National Science Foundation.

\bibliographystyle{apalike}
\bibliography{references}

\end{document}